\documentclass{article}
\usepackage{style/unites}

\usepackage[utf8]{inputenc}
\usepackage[T1]{fontenc}
\usepackage{microtype}

\usepackage{amsmath}
\usepackage{amssymb}
\usepackage{amsfonts}
\usepackage{amsthm}
\usepackage{mathtools}
\usepackage{mathrsfs}
\usepackage{physics}
\usepackage{braket}
\usepackage{slashed}
\usepackage{nicefrac}
\usepackage{textcomp}
\usepackage{dsfont}
\usepackage{bbm}
\usepackage{bm}

\usepackage{graphicx}
\usepackage{subcaption}
\usepackage[export]{adjustbox}
\usepackage{float}
\usepackage{booktabs}
\usepackage{dcolumn}
\newcolumntype{d}[1]{D{.}{.}{#1}}
\usepackage{bigstrut, tabularx, multirow, makecell, diagbox}
\usepackage{colortbl}
\usepackage{tabularray}
\UseTblrLibrary{booktabs}
\usepackage{threeparttable}
\usepackage{tablefootnote}
\usepackage{fontawesome5}

\usepackage{placeins}
\usepackage{caption}
\usepackage{footnote}
\usepackage{enumitem}
\usepackage{multicol}
\usepackage{xspace}
\usepackage{titletoc}
\usepackage{titlesec}
\usepackage[bottom]{footmisc}
\usepackage{setspace}

\usepackage{wrapfig}
\usepackage{tikz}
\usepackage{quantikz}
\usepackage{dashbox}
\usepackage{mdframed}
\usepackage{marvosym}
\usepackage{pifont}
\usepackage{CJK}
\usepackage{url}

\usepackage[table,x11names]{xcolor}
\usepackage[most]{tcolorbox}
\tcbuselibrary{breakable}
\usetikzlibrary{decorations.pathreplacing, fit}

\definecolor{primaryblue}{HTML}{0066CC}
\definecolor{accentcyan}{HTML}{00D4AA}
\definecolor{warmorange}{HTML}{FF6B35}
\definecolor{deepgray}{HTML}{2C3E50}
\definecolor{lightgray}{HTML}{F8F9FA}
\definecolor{gradientstart}{HTML}{667eea}
\definecolor{gradientend}{HTML}{764ba2}

\definecolor{citecolor}{HTML}{0071bc}
\definecolor{citeblue}{RGB}{0, 113, 188}
\definecolor{linkcolor}{HTML}{9A4D92}
\definecolor{firebrick}{rgb}{0.698,0.133,0.133}

\definecolor{paleviolet}{HTML}{E1EEFC}
\definecolor{CarolinaUltraLight}{HTML}{E7F4FC}
\definecolor{lightgrey}{RGB}{247, 247, 247}
\definecolor{shadecolor}{HTML}{EFEFEF}
\definecolor{lightyellow}{rgb}{1.0, 0.95, 0.7}
\definecolor{lightblue}{rgb}{0.90, 0.95, 1.0}
\definecolor{light-gray}{gray}{0.95}

\definecolor{darkgrey}{rgb}{0.5, 0.5, 0.5}
\definecolor{darkgreen}{rgb}{0, 0.5, 0}
\definecolor{mydarkblue}{rgb}{0,0.08,0.45}
\definecolor{mydarkblue2}{rgb}{0.133, 0.133, 0.698}
\definecolor{echodrk}{HTML}{0099cc}
\definecolor{mymauve}{rgb}{0.58,0,0.82}
\definecolor{midnightblue}{rgb}{0.1,0.1,0.44}
\definecolor{oxfordblue}{rgb}{0.0,0.13,0.28}
\definecolor{prussianblue}{rgb}{0.0,0.19,0.33}
\definecolor{coolteal}{rgb}{0, 0.45, 0.45}
\definecolor{olive}{rgb}{0.1, 0.3, 0}
\definecolor{mypurple}{rgb}{0.5,0,0.5}
\definecolor{almond}{rgb}{0.94, 0.87, 0.8}

\definecolor{blue_ampEncoding}{HTML}{DAE8FC}
\definecolor{green_encoder}{HTML}{D5E8D4}
\definecolor{purple_decoder}{HTML}{E1D5E7}
\definecolor{yellow_measure}{HTML}{FFF2CC}
\definecolor{gray_block}{HTML}{F5F5F5}
\definecolor{pink_dru}{HTML}{FAD9D5}
\definecolor{orange_v}{HTML}{FAD7AC}

\definecolor{colorA}{rgb}{1,0,0}
\definecolor{colorB}{rgb}{0,0.3,1}
\definecolor{colorC}{rgb}{0.9,0.8,0.2}
\definecolor{colorD}{rgb}{0,0.65,0}
\definecolor{lesslightgray}{rgb}{0.5,0.5,0.5}
\definecolor{fundamental}{RGB}{55, 110, 111}
\definecolor{Gred}{RGB}{219, 50, 54}
\definecolor{ToCgreen}{RGB}{0, 128, 0}
\definecolor{Sepia}{RGB}{112, 66, 20}
\definecolor{Dblue}{rgb}{0,0.08,0.45}
\definecolor{Blue}{rgb}{0, 0, 0.8}
\definecolor{blue}{rgb}{0,0,1}
\definecolor{UNCblue!10}{rgb}{0.84,0.91,0.98}
\definecolor{RowAlt}{rgb}{0.98,0.98,0.99}

\definecolor{CarolinaBlue}{HTML}{7BAFD4}        
\definecolor{CarolinaLightBlue}{HTML}{B3D4E5}   
\definecolor{CarolinaUltraLight}{HTML}{E8F4F8}  
\definecolor{CarolinaText}{HTML}{1C2B33}        

\usepackage[pagebackref=true,breaklinks=true,colorlinks,hyperfootnotes=false]{hyperref}
\hypersetup{
  colorlinks,
  citecolor=citeblue,
  linkcolor=citeblue,
  urlcolor=citeblue
}
\usepackage[nameinlink,capitalize,noabbrev]{cleveref}

\titlespacing\section{0pt}{4pt plus 4pt minus 2pt}{-2pt plus 2pt minus 2pt}
\titlespacing\subsection{0pt}{2pt plus 4pt minus 2pt}{-2pt plus 2pt minus 2pt}
\titlespacing\subsubsection{0pt}{2pt plus 4pt minus 2pt}{-2pt plus 2pt minus 2pt}

\makeatletter
\def\th@remark{%
  \thm@headfont{\bfseries}%
  \normalfont 
  \thm@preskip\topsep \divide\thm@preskip\tw@
  \thm@postskip\thm@preskip
}
\makeatother

\theoremstyle{definition}

\tcolorboxenvironment{theorem}{
  breakable,
  colback=black!10,
  colframe=white,
  width=\linewidth, 
  enlarge left by=0pt,
  enlarge right by=0pt,
  boxsep=5pt,
  boxrule=0pt,
  left=0pt,right=0pt,top=0pt,bottom=0pt,
  arc=8pt,
  before skip=\topsep,
  after skip=\topsep
}

\tcolorboxenvironment{lemma}{
  breakable,
  colback=black!10,
  colframe=white,
  width=\linewidth,
  enlarge left by=0pt,
  enlarge right by=0pt,
  boxsep=5pt,
  boxrule=0pt,
  left=0pt,right=0pt,top=0pt,bottom=0pt,
  arc=8pt,
  before skip=\topsep,
  after skip=\topsep
}

\tcolorboxenvironment{corollary}{
  breakable,
  colback=black!10,
  colframe=white,
  width=\linewidth,
  enlarge left by=0pt,
  enlarge right by=0pt,
  boxsep=5pt,
  boxrule=0pt,
  left=0pt,right=0pt,top=0pt,bottom=0pt,
  arc=8pt,
  before skip=\topsep,
  after skip=\topsep
}

\tcolorboxenvironment{proposition}{
  breakable,
  colback=black!10,
  colframe=white,
  width=\linewidth,
  enlarge left by=0pt,
  enlarge right by=0pt,
  boxsep=5pt,
  boxrule=0pt,
  left=0pt,right=0pt,top=0pt,bottom=0pt,
  arc=8pt,
  before skip=\topsep,
  after skip=\topsep
}

\tcolorboxenvironment{definition}{
  breakable,
  colback=black!10,
  colframe=white,
  width=\linewidth,
  enlarge left by=0pt,
  enlarge right by=0pt,
  boxsep=5pt,
  boxrule=0pt,
  left=0pt,right=0pt,top=0pt,bottom=0pt,
  arc=8pt,
  before skip=\topsep,
  after skip=\topsep
}

\tcolorboxenvironment{assumption}{
  breakable,
  colback=black!10,
  colframe=white,
  width=\linewidth,
  enlarge left by=0pt,
  enlarge right by=0pt,
  boxsep=5pt,
  boxrule=0pt,
  left=0pt,right=0pt,top=0pt,bottom=0pt,
  arc=8pt,
  before skip=\topsep,
  after skip=\topsep
}

\tcolorboxenvironment{claim}{
  breakable,
  colback=black!10,
  colframe=white,
  width=\linewidth,
  enlarge left by=0pt,
  enlarge right by=0pt,
  boxsep=5pt,
  boxrule=0pt,
  left=0pt,right=0pt,top=0pt,bottom=0pt,
  arc=8pt,
  before skip=\topsep,
  after skip=\topsep
}

\tcolorboxenvironment{problem}{
  breakable,
  colback=black!10,
  colframe=white,
  width=\linewidth,
  enlarge left by=0pt,
  enlarge right by=0pt,
  boxsep=5pt,
  boxrule=0pt,
  left=0pt,right=0pt,top=0pt,bottom=0pt,
  arc=8pt,
  before skip=\topsep,
  after skip=\topsep
}

\tcolorboxenvironment{question}{
  breakable,
  colback=black!10,
  colframe=white,
  width=\linewidth,
  enlarge left by=0pt,
  enlarge right by=0pt,
  boxsep=5pt,
  boxrule=0pt,
  left=0pt,right=0pt,top=0pt,bottom=0pt,
  arc=8pt,
  before skip=\topsep,
  after skip=\topsep
}

\newtcolorbox{titleblock}{
  enhanced,
  frame hidden,
  colback=CarolinaUltraLight,
  colframe=CarolinaUltraLight,
  boxrule=0pt,
  arc=10pt,
  left=14pt,
  right=14pt,
  top=14pt,
  bottom=14pt,
  width=\linewidth,
  before skip=12pt plus 4pt,
  after skip=12pt plus 4pt,
  grow to left by=1.5pt,
  grow to right by=1.5pt,
  before upper={
    \setlength{\parindent}{0cm}
    \setlength{\parskip}{0.5cm}
  }
}

\crefname{theorem}{Theorem}{Theorems}
\crefname{proposition}{Proposition}{Propositions}
\crefname{lemma}{Lemma}{Lemmas}
\crefname{corollary}{Corollary}{Corollaries}
\crefname{definition}{Definition}{Definitions}
\crefname{assumption}{Assumption}{Assumptions}
\crefname{remark}{Remark}{Remarks}
\crefname{problem}{Problem}{Problems}
\crefname{property}{Property}{property}
\crefname{question}{Question}{Questions}

\numberwithin{equation}{section}
\numberwithin{theorem}{section}
\numberwithin{proposition}{section}
\numberwithin{definition}{section}
\numberwithin{lemma}{section}
\numberwithin{assumption}{section}
\numberwithin{remark}{section}

\newcommand\metadataformat[2][]{{\small {\bfseries #1:} #2}}

\def\1{\bm{1}}

\makeatletter
\let\save@mathaccent\mathaccent
\newcommand*\if@single[3]{%
    \setbox0\hbox{${\mathaccent"0362{#1}}^H$}%
    \setbox2\hbox{${\mathaccent"0362{\kern0pt#1}}^H$}%
    \ifdim\ht0=\ht2 #3\else #2\fi
}
\newcommand*\rel@kern[1]{\kern#1\dimexpr\macc@kerna}
\newcommand*\widebar[1]{\@ifnextchar^{{\wide@bar{#1}{0}}}{\wide@bar{#1}{1}}}
\newcommand*\wide@bar[2]{\if@single{#1}{\wide@bar@{#1}{#2}{1}}{\wide@bar@{#1}{#2}{2}}}
\newcommand*\wide@bar@[3]{%
    \begingroup
    \def\mathaccent##1##2{%
        \let\mathaccent\save@mathaccent
        \if#32 \let\macc@nucleus\first@char \fi
        \setbox\z@\hbox{$\macc@style{\macc@nucleus}_{}$}%
        \setbox\tw@\hbox{$\macc@style{\macc@nucleus}{}_{}$}%
        \dimen@\wd\tw@
        \advance\dimen@-\wd\z@
        \divide\dimen@ 3
        \@tempdima\wd\tw@
        \advance\@tempdima-\scriptspace
        \divide\@tempdima 10
        \advance\dimen@-\@tempdima
        \ifdim\dimen@>\z@ \dimen@0pt\fi
        \rel@kern{0.6}\kern-\dimen@
        \if#31
        \overline{\rel@kern{-0.6}\kern\dimen@\macc@nucleus\rel@kern{0.4}\kern\dimen@}%
        \advance\dimen@0.4\dimexpr\macc@kerna
        \let\final@kern#2%
        \ifdim\dimen@<\z@ \let\final@kern1\fi
        \if\final@kern1 \kern-\dimen@\fi
        \else
        \overline{\rel@kern{-0.6}\kern\dimen@#1}%
        \fi
    }%
    \macc@depth\@ne
    \let\math@bgroup\@empty \let\math@egroup\macc@set@skewchar
    \mathsurround\z@ \frozen@everymath{\mathgroup\macc@group\relax}%
    \macc@set@skewchar\relax
    \let\mathaccentV\macc@nested@a
    \if#31
    \macc@nested@a\relax111{#1}%
    \else
    \def\gobble@till@marker##1\endmarker{}%
    \futurelet\first@char\gobble@till@marker#1\endmarker
    \ifcat\noexpand\first@char A\else
    \def\first@char{}%
    \fi
    \macc@nested@a\relax111{\first@char}%
    \fi
    \endgroup
    }
\makeatother

\DeclareMathAlphabet{\mathsfit}{\encodingdefault}{\sfdefault}{m}{sl}
\SetMathAlphabet{\mathsfit}{bold}{\encodingdefault}{\sfdefault}{bx}{n}

\let\tilde\widetilde

\usepackage{ulem}
\usepackage{algorithm}       
\usepackage{algpseudocode}   

\usepackage{listings}

\newcommand{\appcontents}{%
  \clearpage
  \phantomsection
  \startcontents[app]
  \printcontents[app]{l}{1}{\section*{Appendix Contents}}%
}

\begin{document}

\makeatletter
\def\blfootnote{\gdef\@thefnmark{}\@footnotetext}
\makeatother

\makeatletter
\pagestyle{fancy}
\fancyhf{}
\renewcommand{\headrulewidth}{1pt}
\chead{\small\bf \texttt{FlashSchNet}: Fast and Accurate Coarse-Grained Neural Network Molecular Dynamics
}
\cfoot{\thepage}
\thispagestyle{fancy}
\makeatother

\makeatletter
\def\icmldate#1{\gdef\@icmldate{#1}}
\icmldate{\today}
\makeatother

\makeatletter
\fancypagestyle{fancytitlepage}{
  \fancyhead{}
  \lhead{\includegraphics[height=0.8cm]{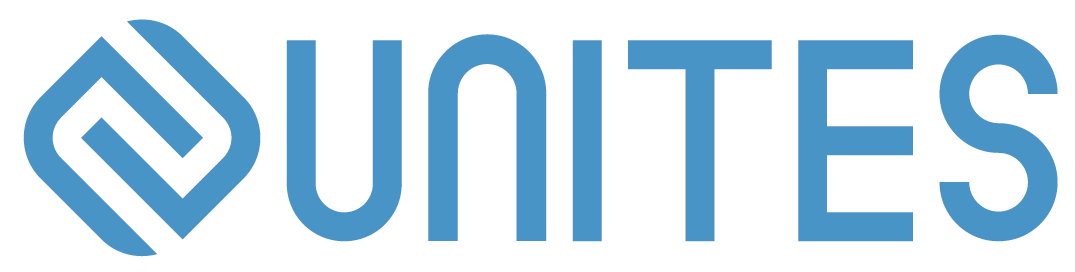}}
  \rhead{\it \@icmldate}
  \cfoot{}
}
\makeatother

\thispagestyle{fancytitlepage}

\vspace*{0.5em}

\noindent
\begin{titleblock}
    {\setlength{\parskip}{0cm}
     \raggedright
     {\setstretch{1.2}
      \LARGE\sffamily\bfseries
      \texttt{FlashSchNet}: Fast and Accurate Coarse-Grained Neural Network Molecular Dynamics
      \par}
    }
    \vskip 0.2cm
    
    \begin{icmlauthorlist}
\mbox{Pingzhi Li$^{1}$},
\mbox{Hongxuan Li$^{1,2}$},
\mbox{Zirui Liu$^{3}$},
\mbox{Xingcheng Lin$^{2}$},
and \mbox{Tianlong Chen$^{1}$}
\end{icmlauthorlist}

$^{1\,}$UNC-Chapel Hill
\quad $^{2\,}$North Carolina State University
\quad $^{3\,}$University of Minnesota Twin Cities
    \vskip 0.2cm
    
    Graph neural network~(GNN) potentials such as SchNet improve the accuracy and transferability of molecular dynamics~(MD) simulation by learning many-body interactions, but remain slower than classical force fields due to fragmented kernels and memory-bound pipelines that underutilize GPUs.
We show that a missing principle is making GNN-MD \emph{IO-aware}, carefully accounting for reads and writes between GPU high-bandwidth memory (HBM) and on-chip SRAM.
We present \texttt{FlashSchNet}, an efficient and accurate IO-aware SchNet-style GNN-MD framework built on four techniques:
(1)~\emph{flash radial basis}, which fuses pairwise distance computation, Gaussian basis expansion, and cosine envelope into a single tiled pass, computing each distance once and reusing it across all basis functions;
(2)~\emph{flash message passing}, which fuses cutoff, neighbor gather, filter multiplication, and reduction to avoid materializing edge tensors in HBM;
(3)~\emph{flash aggregation}, which reformulates scatter-add via CSR segment reduce, reducing atomic writes by a factor of feature dimension and enabling contention-free accumulation in both forward and backward passes;
(4)~\textit{channel-wise 16-bit quantization} that exploits the low per-channel dynamic range in SchNet MLP weights to further improve throughput with negligible accuracy loss.
On a single NVIDIA RTX PRO 6000, \texttt{FlashSchNet} achieves \textbf{1000~ns/day} aggregate simulation throughput over 64 parallel replicas on coarse-grained~(CG) protein containing 269 beads (\textbf{6.5}$\mathbf{\times}$ faster than CGSchNet baseline with \textbf{80\% reduction} of peak memory), surpassing classical force fields (\emph{e.g.} MARTINI) while retaining SchNet-level accuracy and transferability.
    
    \vskip 0.2cm
    {\setlength{\parskip}{0cm}
     \centering
     \makebox[\linewidth]{
        \metadataformat[Code]{
            \href{https://github.com/UNITES-Lab/flash-molecular-dynamics}{https://github.com/UNITES-Lab/flash-molecular-dynamics}
        }
     }
    }
\end{titleblock}

\blfootnote{%
$^{\textrm{\Letter}}$ Correspondence email: \{pingzhi, tianlong\}@cs.unc.edu, xingcheng\_lin@ncsu.edu, zrliu@umn.edu
\\[2.5em]
\ifcsname @icmlpreprint\endcsname
  \textit{\csname @icmlpreprint\endcsname}%
\fi
}

\section{Introduction}
\label{sec:intro}
\begin{figure}[htbp]
\centering
\includegraphics[width=0.75\linewidth]{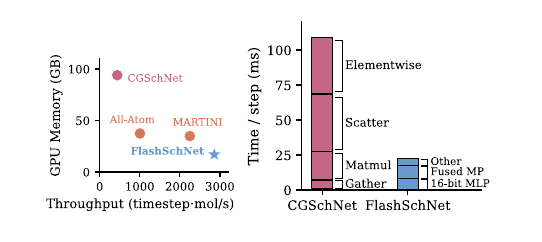}
\caption{\uline{Left:} Memory-throughput trade-off for SchNet-style GNN-MD. \texttt{FlashSchNet} achieves $5\times$ memory reduction while improving throughput by $6\times$ over CGSchNet baseline. \uline{Right:} Step time breakdown showing \texttt{FlashSchNet} eliminates scatter and element-wise bottlenecks via fused kernels and 16-bit quantization. All are evaluated on a 269-bead protein (1ENH) with 64 replicas.}
\label{fig:teaser-breakdown}
\end{figure}
Molecular dynamics~(MD) simulation is a core tool in computational chemistry, drug discovery, and materials science, offering a computational microscope for probing molecular motion at atomic resolution~\citep{karplus2002molecular,dror2012biomolecular,hollingsworth2018molecular}.
By numerically integrating Newton's equations of motion, MD produces time-resolved trajectories that connect microscopic interactions to macroscopic observables, enabling thermodynamic estimation, conformational exploration, and mechanistic study of rare events~\citep{chodera2014markov}.
In practice, however, classical MD faces a persistent tradeoff: empirical force fields are fast but approximate, while first-principles MD such as Car--Parrinello is more faithful but orders of magnitude more expensive~\citep{car1985unified}.
Even when adopting widely used coarse-grained models such as MARTINI~\citep{marrink2007martini}, which sacrifice atomistic detail for efficiency, the repeated evaluation of forces over millions to billions of timesteps remains a fundamental bottleneck, limiting accessible timescales and system sizes in routine workflows~\citep{de2016role}.

Motivated by this gap, graph neural network~(GNN) potentials have rapidly emerged as a leading class of machine-learned force fields (MLFFs).
Rooted in geometric deep learning principles~\citep{bronstein2021geometric}, these models represent atoms as nodes and local interactions as edges, and use message passing to capture many-body effects in a data-driven yet physically structured manner~\citep{gilmer2017neural}.
SchNet~\citep{schutt2017schnet} and subsequent geometric GNNs (for example, DimeNet~\citep{gasteiger2020directional} and $E(3)$-equivariant architectures such as NequIP~\citep{batzner20223}, Allegro~\citep{musaelian2023learning}, and MACE~\citep{batatia2022mace}) have demonstrated strong accuracy and improved transferability across chemical environments, bringing ML potentials closer to first-principles fidelity at a fraction of the compute.

Yet higher accuracy has not translated into faster wall-clock simulation.
In SchNet-style GNN-MD, continuous-filter convolution~(CFConv) repeatedly constructs edge-wise features (distances, radial bases, cutoffs) and applies small MLPs, followed by scatter-style aggregation over dynamic neighborhoods.
When implemented with high-level deep learning frameworks such as PyTorch and JAX, this computation fragments into many kernels and repeatedly materializes intermediate edge tensors in GPU high-bandwidth memory~(HBM), while aggregation suffers from heavy synchronization due to contended atomic updates.
As a result, the workload is strongly memory-bound and underutilizes the GPU despite modest nominal FLOPs. For example, CGSchNet~\citep{charron2025navigating} running 64 parallel replicas on a coarse-grained~(CG) protein with 269 beads achieves only \textbf{2.5\%} model FLOPs utilization~(MFU)\footnote{MFU is computed as achieved TFLOPs/s divided by the GPU peak TFLOPs/s. We enable TF32 tensor cores and report measurements on a single NVIDIA RTX PRO 6000.}.
These observations point to a missing principle for practical GNN-MD: \textit{making the pipeline IO-aware by optimizing reads and writes between HBM and on-chip SRAM.}

We identify four major bottlenecks in SchNet-style GNN-MD, which all stem from memory IO:
\ding{182}~\emph{Radial basis expansion} computes pairwise distances, Gaussian basis values, and cosine cutoffs in separate kernels, materializing intermediate tensors (distances, expanded bases, cutoff values) to HBM even though each is consumed only once;
\ding{183}~\emph{Message passing} launches distinct operations for cutoff masking, neighbor gather, filter multiplication, and scatter aggregation, writing large edge tensors of size $O(E \times F)$ (number of edges $\times$ feature dimension) to HBM between stages;
\ding{184}~\emph{Scatter aggregation} uses atomic additions to accumulate messages, incurring $O(E \times F)$ conflict atomic writes that serialize under high neighborhood density;
\ding{185}~\emph{Filter networks} repeatedly load MLP weights for every edge, making these small matrix multiplications strongly bandwidth-bound.

We propose: \ding{182} \emph{Flash radial basis} fuses pairwise distance computation, Gaussian basis expansion, and cosine envelope into a single tiled pass, computing each distance once and reusing it on-chip across all basis functions.
\ding{183} \emph{Flash message passing} fuses cutoff masking, neighbor gather, filter multiplication, and reduction into one kernel, eliminating materialization of intermediate edge tensors in HBM.
\ding{184} \emph{Flash aggregation} reformulates scatter-add via CSR segment reduce, reducing atomic writes by a factor of feature dimension and enabling contention-free accumulation in both forward and backward passes.
\ding{185} \emph{Channel-wise 16-bit quantization} exploits low per-channel dynamic range in SchNet MLP weights to further improve throughput with negligible accuracy loss. As demonstrated in Figure~\ref{fig:teaser-breakdown}, \texttt{FlashSchNet} achieves significant end-to-end speedup and memory savings. Our contributions are summarized as:
\begin{itemize}
    \item We identify IO inefficiency as the key bottleneck in SchNet-style GNN-MD and show how to exploit inherent model structure, \textit{i.e.} graph sparsity for contention-free CSR aggregation and low per-channel dynamic range for 16-bit weight quantization, to reduce memory traffic at the algorithmic level.

    \item We translate these insights into four kernel-level implementations (\textit{i.e.}, flash radial basis, flash message passing, flash aggregation, and quantized filter networks) that together eliminate intermediate tensor materialization, deliver end-to-end speedup and memory saving.

    \item We combine these techniques into \texttt{FlashSchNet}, achieving \textbf{6.5$\times$ speedup} and \textbf{80\% memory reduction} over CGSchNet baseline. To our knowledge, this is the first SchNet-style GNN-MD that surpasses classical coarse-grained force fields such as MARTINI, reaching \textbf{1000~ns/day} aggregate simulation throughput over 64 parallel replicas on coarse-grained protein containing 269 beads on a single RTX PRO 6000, while retaining the accuracy and transferability of learned potentials.
\end{itemize}
\section{Related Work}
\label{sec:related_work}
\paragraph{Molecular dynamics simulation and machine-learned force fields.}
Molecular dynamics~(MD) simulation is fundamental for studying molecular systems. Traditional force fields (e.g., AMBER~\cite{wang2004development}, CHARMM~\cite{brooks2009charmm}) are computationally efficient but limited in transferability due to fixed functional forms. \emph{Ab initio} MD~\cite{car1985unified, kuhne2020cp2k} achieves high accuracy via first-principles calculations, but its $\mathcal{O}(N^3)$ scaling restricts applicability.
Machine-learned force fields~(MLFFs) bridge this accuracy-efficiency gap. Early approaches include kernel methods, \textit{e.g.} GAP~\cite{bartok2010gaussian} and sGDML~\cite{chmiela2019sgdml}, and neural network potentials~\cite{behler2007generalized}. GNN-based MLFFs such as SchNet~\cite{schutt2017schnet}, DimeNet~\cite{gasteiger2020directional}, and PhysNet~\cite{unke2019physnet} operate directly on molecular graphs with improved generalization. $E(3)$-equivariant models including NequIP~\cite{batzner20223}, Allegro~\cite{musaelian2023learning}, and MACE~\cite{batatia2022mace} achieve superior data efficiency by preserving geometric symmetries, though at higher computational cost. Recent universal MLFFs~\cite{ju2025application,neumann2024orb,yang2024mattersim} enhance transferability through large-scale training, but inference cost remains a bottleneck for large-scale simulations.

Recent works have explored coarse-grained GNN force fields to improve scalability while maintaining physical fidelity.
Airas and Zhang~\cite{airas2026knowledge} introduce a solvent-aware CG potential by distilling structural priors from protein language models, focusing on secondary structure and solvent exposure.
Majewski et al.~\cite{majewski2023machine} develop neural CG force fields that reproduce protein thermodynamics across multiple proteins using long atomistic trajectories.
Charron et al.~\cite{charron2025navigating} propose a transferable CG model that generalizes to unseen sequences and accurately predicts folding landscapes and mutation effects.  While these models improve physical fidelity and, in some cases, generalization, their inference remains memory- and IO-bound, limiting scalability to long trajectories or large biomolecular systems.

\paragraph{Efficient graph neural networks.}
A line of work improves GNN efficiency by optimizing sparse message-passing operators at various system levels, including graph-centric frameworks with dedicated CUDA kernels~\citep{wang2019deep,fey2019fast}, runtime systems that adapt execution to graph structure~\citep{wang2021gnnadvisor}, compiler stacks that fuse operators and reduce kernel launches~\citep{xie2022graphiler}, and accelerated sparse primitives such as SpMM and SDDMM with CSR-compatible designs~\citep{chen2020fusegnn,huang2020ge,rahman2021fusedmm}.
These efforts primarily target generic GNN workloads on large, mostly static graphs where sparse linear algebra dominates.
SchNet-style GNN-MD differs significantly because dynamic neighbor lists, continuous-filter convolutions with per-edge MLPs, and the need for efficient backward passes for force computation make repeated edge-tensor materialization and contention-heavy scatter-add the key bottlenecks, motivating our IO-aware fusion and contention-free CSR-style aggregation.

\paragraph{Memory-bound runtime optimization.}
Modern GNN and ML potential workloads are often memory-bound, as irregular gather/scatter interleaved with small dense kernels makes throughput dominated by data movement rather than FLOPs.
FlashAttention~\citep{dao2022flashattention} exemplifies IO-aware algorithm design that explicitly reasons about HBM to SRAM traffic and uses tiling and recomputation to maximize on-chip reuse.
In GNN runtimes, high-level frameworks often execute message construction and aggregation as fragmented kernels that materialize intermediates and suffer from atomic contention~\citep{gong2025identifying}, prompting fusion techniques that reduce memory traffic and launch overhead~\citep{liu2024df}.
For molecular simulation, TorchMD-Net 2.0 achieves substantial speedups by engineering the simulation stack with optimized neighbor search and efficient force evaluation~\citep{pelaez2024torchmd}.
These efforts highlight that large wall-clock gains require end-to-end, IO-aware redesign that co-optimizes feature construction, message passing, and aggregation.
\section{Background}\label{sec:background}

We summarize our used notation in Table~\ref{tab:notation}. Section~\ref{sec:md-force} reviews molecular dynamics and force evaluation. Section~\ref{sec:schnet-model} presents the SchNet architecture, highlighting the operators that dominate runtime. Section~\ref{sec:challenge-hw} demonstrates the hardware bottlenecks that motivate our IO-aware design.

\begin{table}[t]
\centering
\caption{Summary of notation used throughout the paper.}
\label{tab:notation}
\resizebox{0.65\linewidth}{!}{
\begin{tabular}{l|l}
\toprule[1pt]
\midrule
Symbol & Description \\
\midrule
$N$ & Number of atoms or beads \\
$E$ & Number of directed edges in the neighbor graph \\
$D$ & Hidden feature dimension \\
$D_r$ & Radial basis dimension \\
$T$ & Number of interaction blocks \\
$r_{\text{cut}}$ & Cutoff radius for neighbor list construction \\
\midrule
$\mathbf{r}_i \in \mathbb{R}^3$ & Position of atom $i$ \\
$\mathbf{x}_i^{(t)} \in \mathbb{R}^D$ & Hidden feature of atom $i$ at layer $t$ \\
$\mathtt{X}^{(t)} \in \mathbb{R}^{N \times D}$ & Stacked hidden features over all atoms \\
\midrule
$\mathtt{src}, \mathtt{dst} \in \{1,\dots,N\}^E$ & Source and destination index arrays for edges \\
$d_e \in \mathbb{R}$ & Scalar distance for edge $e$ \\
$\mathbf{b}_e \in \mathbb{R}^{D_r}$ & Radial basis vector for edge $e$ \\
$\mathtt{B} \in \mathbb{R}^{E \times D_r}$ & Stacked radial basis over all edges \\
$\mathbf{w}_e \in \mathbb{R}^D$ & Continuous filter for edge $e$ \\
$\mathtt{W} \in \mathbb{R}^{E \times D}$ & Stacked filters over all edges \\
$\mathbf{m}_e^{(t)} \in \mathbb{R}^D$ & Message for edge $e$ at layer $t$ \\
\midrule
$\mathcal{E}$ & Total potential energy \\
$\epsilon_i \in \mathbb{R}$ & Per-atom energy contribution from atom $i$ \\
$\mathbf{F}_i \in \mathbb{R}^3$ & Force on atom $i$ \\
\midrule
\bottomrule[1pt]
\end{tabular}
}
\end{table}

\subsection{Molecular dynamics and force evaluation}\label{sec:md-force}
Molecular dynamics (MD) simulates the evolution of atom/bead positions $\{\mathbf{r}_i\}_{i=1}^{N}$ by repeatedly evaluating forces $\{\mathbf{F}_i\}$ and integrating the equations of motion~\citep{karplus2002molecular}.
In energy-based MD, a force field defines a scalar potential energy $\mathcal{E}(\{\mathbf{r}_i\})$, and forces are:
\begin{equation*}
\mathbf{F}_i = -\nabla_{\mathbf{r}_i}\mathcal{E} \in \mathbb{R}^3.
\end{equation*}
A time integrator then updates the state via:
\begin{equation*}
\mathbf{r}_i \leftarrow \mathbf{r}_i + \Delta t \, \mathbf{v}_i + \cdots,
\qquad
\mathbf{v}_i \leftarrow \mathbf{v}_i + \Delta t \, \mathbf{F}_i/m_i + \cdots,
\end{equation*}
where $\mathbf{v}_i$ is the velocity, $m_i$ is the mass, and the omitted terms depend on the chosen thermostat integrator (\textit{e.g.}, Langevin).
Each MD step therefore requires (i) evaluating $\mathcal{E}$ and (ii) backpropagating to obtain $\mathbf{F}_i$, making the end-to-end throughput dominated by both \emph{forward} and \emph{backward} cost of the learned potential.

\subsection{SchNet model}\label{sec:schnet-model}
SchNet~\citep{schutt2017schnet} is a continuous-filter message-passing network that predicts potential energy $E$ from atom positions $\{\mathbf{r}_i\}_{i=1}^{N}$ and types $\{Z_i\}_{i=1}^{N}$, and obtains forces via $\mathbf{F}_i=-\nabla_{\mathbf{r}_i}E$.
The model maintains atom-wise hidden features $\mathbf{x}_i^{(t)}\in\mathbb{R}^{D}$ (stacked as $\mathtt{X}^{(t)}\in\mathbb{R}^{N\times D}$) and iteratively applies distance-dependent interactions over a sparse neighbor graph induced by a radial cutoff.
We now describe the computational pipeline and dominating operators, following the specific architecture used in \citet{charron2025navigating}.

\paragraph{Building the neighbor list.}
Given positions, SchNet first constructs a neighbor list with cutoff radius $r_{\text{cut}}$, represented as two index arrays $\mathtt{src},\mathtt{dst}\in\{1,\dots,N\}^{E}$ indexing the $E$ directed edges.
Each edge $e$ encodes an interaction from source $j=\mathtt{src}[e]$ to destination $i=\mathtt{dst}[e]$.

\paragraph{Distances and radial basis.}
For each edge $e$, SchNet computes the displacement vector and scalar distance
\begin{equation*}
\mathbf{u}_e=\mathbf{r}_{\mathtt{dst}[e]}-\mathbf{r}_{\mathtt{src}[e]}\in\mathbb{R}^{3},\qquad
d_e=\|\mathbf{u}_e\|_2,
\end{equation*}
and expands $d_e$ into a $D_r$-dimensional radial basis vector $\mathbf{b}_e=\mathrm{RBF}(d_e)\in\mathbb{R}^{D_r}$, typically modulated by a smooth cutoff envelope.
Stacking over all edges yields $\mathtt{B}\in\mathbb{R}^{E\times D_r}$.

\paragraph{Filter network.}
A small MLP maps each radial basis vector to a $D$-dimensional continuous filter:
\begin{equation*}
\mathbf{w}_e=\mathrm{MLP}_{\text{filter}}(\mathbf{b}_e)\in\mathbb{R}^{D},
\end{equation*}
producing stacked filters $\mathtt{W}\in\mathbb{R}^{E\times D}$.
Because this MLP is evaluated per edge, the resulting tensor scales as $O(E \times D)$ and constitutes a major source of memory traffic.

\paragraph{CFConv message passing and aggregation.}
The continuous-filter convolution (CFConv) forms edge messages by element-wise multiplication of the source feature with the learned filter,
$\mathbf{m}_e^{(t)}=\mathbf{x}_{\mathtt{src}[e]}^{(t)}\odot \mathbf{w}_e\in\mathbb{R}^{D}$,
and aggregates them onto destination nodes via a sum over incoming edges:
$\mathbf{h}_i^{(t)}=\sum_{e:\ \mathtt{dst}[e]=i}\mathbf{m}_e^{(t)}\in\mathbb{R}^{D}$.
A point-wise update network with residual connection then produces $\mathbf{x}_i^{(t+1)}$; this interaction block is repeated $T$ times.
Standard implementations realize aggregation via \texttt{scatter\_add}, which incurs $O(E \times D)$ atomic writes with significant contention when multiple edges share the same destination.

\paragraph{Energy readout.}
After $T$ interaction blocks, an output MLP maps atom features to per-atom energy contributions:
\begin{equation*}
\epsilon_i = \mathrm{MLP}_{\text{out}}(\mathbf{x}_i^{(T)}) \in \mathbb{R},\qquad
E = \sum_{i=1}^{N} \epsilon_i,
\end{equation*}
optionally combined with prior energy terms (\textit{e.g.}, bonded interactions) as $E \leftarrow E + E_{\text{prior}}$~\citep{charron2025navigating}.
SchNet contains multiple MLP submodules, \textit{i.e.}~$\mathrm{MLP}_{\text{filter}}$, block-wise update networks, and $\mathrm{MLP}_{\text{out}}$. All of them are bandwidth-bound due to repeated weight loading.

\paragraph{Forces via autodiff.}
For molecular dynamics, forces are obtained by differentiating the scalar energy with respect to positions:
$\mathbf{F}_i = -\nabla_{\mathbf{r}_i} E \in \mathbb{R}^3$.
This requires backpropagating through neighbor-list-indexed distance and RBF computations, all MLP submodules, and the aggregation operator.
Crucially, the backward pass through aggregation also involves scatter-style accumulation (now over source nodes), making both forward and backward efficiency essential for practical MD throughput.

\subsection{Challenges of hardware performance}\label{sec:challenge-hw}
We focus on GPU. Modern GPUs offer high peak FLOPs but are frequently limited by memory traffic between high-bandwidth memory (HBM) and on-chip storage.
For SchNet-style GNN-MD, the dominant operators are sparse, index-based pipelines over the neighbor graph, as they repeatedly materialize large edge tensors (\textit{e.g}., $\mathtt{B}\in\mathbb{R}^{E\times D_r}$ and $\mathtt{W}\in\mathbb{R}^{E\times D}$) and perform scatter-style reductions whose arithmetic intensity is low relative to their HBM read/write volume.
This makes runtime primarily \emph{bandwidth-bound}, and fragmented kernels further reduce effective throughput by repeatedly loading and storing intermediates.

\paragraph{HBM traffic from edge intermediates.}
The core SchNet pipeline expands edge distances into $\mathtt{B}\in\mathbb{R}^{E\times D_r}$ and filters into $\mathtt{W}\in\mathbb{R}^{E\times D}$, and conceptually forms edge messages $\mathtt{M}^{(t)}\in\mathbb{R}^{E\times D}$ with rows $\mathbf{m}_e^{(t)}$.
Even when compute per element is modest, writing and rereading these edge tensors incurs $O(E\cdot D_r)$ and $O(E\cdot D)$ HBM traffic per block, which is amplified across $T$ interaction blocks and again in the backward pass required for force evaluation.

\paragraph{Scatter contention in graph reductions.}
Aggregation in CFConv typically uses \texttt{scatter\_add} to accumulate $\mathbf{m}_e^{(t)}$ into destination nodes $\mathbf{h}_i^{(t)}$.
This performs $O(E\cdot D)$ atomic updates, and when many edges share the same destination (\textit{i.e.}, high local degree), concurrent atomics serialize and significantly lower throughput.
Moreover, the backward pass through aggregation also requires scatter accumulation, so contention impacts both forward and backward passes, directly limiting MD wall-clock step time.

\begin{figure}[t]
    \centering
    \includegraphics[width=0.95\linewidth]{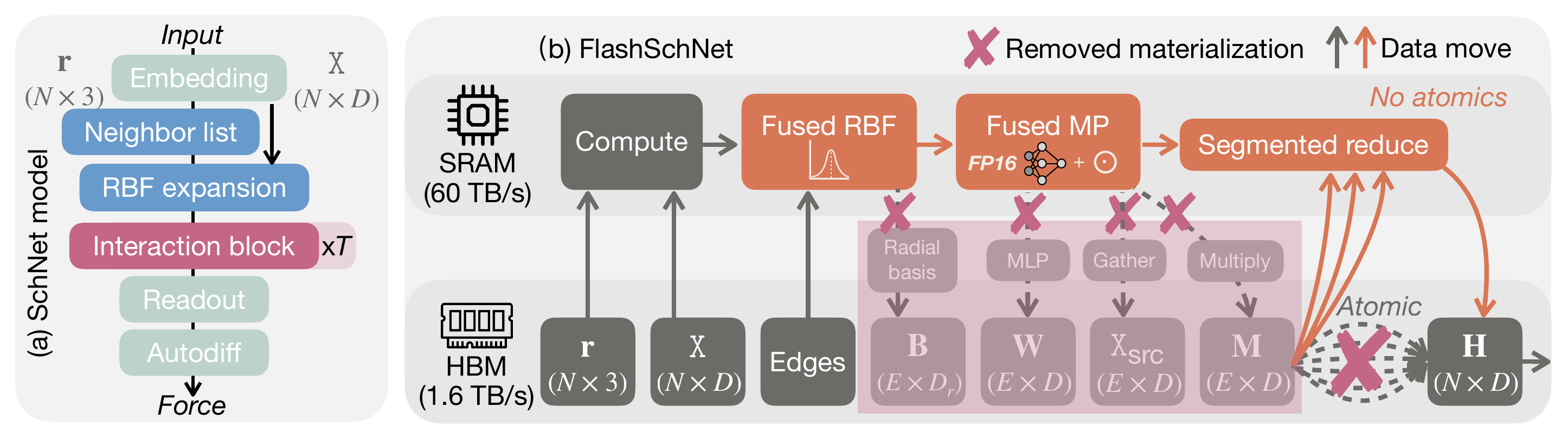}
    \caption{\uline{(a)} SchNet model architecture for molecular dynamics: atom positions $\mathbf{r}$ and embeddings $\mathbf{X}$ are processed through neighbor list construction, radial basis expansion, and $T$ interaction blocks, followed by energy readout and autodiff for force computation. \uline{(b)} \texttt{FlashSchNet} IO-aware execution model. The baseline pipeline (bottom, shaded) materializes intermediate edge tensors ($\mathbf{B} \in \mathbb{R}^{E \times D_r}$, $\mathbf{W}, \mathbf{X}_{\text{src}}, \mathbf{M} \in \mathbb{R}^{E \times D}$) to HBM and uses atomic scatter for aggregation. \texttt{FlashSchNet} (top, orange) fuses these operations into three kernels that keep intermediates in SRAM: \textit{Fused RBF} computes distances, Gaussian basis expansion, and cosine envelope in one pass; \textit{Fused MP} combines FP16 filter MLP, neighbor gather, and element-wise multiplication; \textit{Segmented reduce} replaces atomic scatter-add with contention-free CSR-style accumulation. Red crosses indicate eliminated HBM materializations. The \texttt{FlashSchNet} pipeline reduces memory traffic by ${\sim}E/N$ and removes all atomic contention.}
    \label{fig:method}
\end{figure}

\paragraph{Mixed precision and Tensor Cores.}
GPUs provide specialized Tensor Cores that accelerate matrix-multiply and fused MLP primitives at various precisions (\textit{e.g.} FP16).
In SchNet, the filter, update, and readout networks are composed of MLPs whose weights are repeatedly loaded, making them sensitive to both compute throughput and memory bandwidth.
Using FP16 weights and activations, while keeping key accumulations and force outputs in FP32, can reduce weight/activation traffic and increase compute throughput by mapping these MLPs onto Tensor Cores.

\section{\texttt{Flash-SchNet}}
\label{sec:flash_schnet}

This section presents \texttt{FlashSchNet}, an IO-aware SchNet-style GNN-MD implementation that accelerates the \emph{end-to-end} MD step, including forward energy evaluation + backward force computation.
As shown in Figure~\ref{fig:method}, our design targets the dominant bottlenecks identified in Section~\ref{sec:background}.
At a high level, \texttt{FlashSchNet} follows the baseline CGSchNet~\citep{charron2025navigating} architecture, but (a) fuses single-use edge pipelines to avoid materializing large intermediates in HBM, (b) replaces atomic scatter reductions with contention-free CSR segment reductions, and (c) applies channel-wise 16-bit quantization to MLP submodules to reduce both compute time and memory traffic losslessly.

\paragraph{Computation targeted by \texttt{FlashSchNet}.}
Consider one interaction block at layer index $t$ under neighbor graph $(\mathtt{src},\mathtt{dst})\in\{1,\dots,N\}^E$.
For edge $e$, define displacement $\mathbf{u}_e=\mathbf{r}_{\mathtt{dst}[e]}-\mathbf{r}_{\mathtt{src}[e]}$, distance $d_e=\|\mathbf{u}_e\|_2$, radial basis $\mathbf{b}_e=\mathrm{RBF}(d_e)\in\mathbb{R}^{D_r}$, and a smooth cutoff envelope $C(d_e)\in\mathbb{R}$.
The CFConv aggregation can be written as
\begin{align*}
\mathbf{h}_i^{(t)}=\sum_{e:\ \mathtt{dst}[e]=i}\Big(\mathbf{x}_{\mathtt{src}[e]}^{(t)}\odot \mathbf{w}_e\Big),
\end{align*}
where $\mathbf{w}_e=\mathrm{MLP}_{\mathrm{filter}}\!\big(\mathbf{b}_e\cdot C(d_e)\big)\in\mathbb{R}^{D}$. Baseline implementations typically materialize $\mathtt{B}\in\mathbb{R}^{E\times D_r}$ and $\mathtt{W}\in\mathbb{R}^{E\times D}$ as HBM intermediates, and realize the sum using \texttt{scatter\_add}, leading to large memory traffic and atomic contention.
\texttt{FlashSchNet} computes the same $\mathbf{h}_i^{(t)}$ while avoiding edge tensor materialization and eliminating atomics on the aggregation path.

\subsection{IO-aware reformulation of SchNet interaction}\label{sec:io-aware-reform}

\paragraph{Single-use edge pipeline.}
The per-edge computation forms a single-use chain
distance to radial-basis to filter MLP to gated message.
We treat this chain as a streaming operator and fuse it so that $\mathbf{u}_e$, $d_e$, $\mathbf{b}_e$, and intermediate activations inside $\mathrm{MLP}_{\mathrm{filter}}$ are produced and consumed on chip.
Conceptually, we replace explicit edge tensors with a fused edge operator
$\mathbf{h}_i^{(t)}=\sum_{e:\ \mathtt{dst}[e]=i}\Psi\!\big(\mathbf{x}_{\mathtt{src}[e]}^{(t)},\mathbf{r}_{\mathtt{src}[e]},\mathbf{r}_{\mathtt{dst}[e]}\big)$,
where $\Psi$ encapsulates distance computation, radial basis and envelope evaluation, filter MLP, and gating.

\paragraph{Precision contract for force-based simulation.}
Forces require gradients through $d_e=\|\mathbf{u}_e\|_2$ and the cutoff and basis functions.
We keep positions $\mathbf{r}_i$, distances $d_e$, energy accumulation $\mathcal{E}$, and force outputs $\mathbf{F}_i$ in FP32, and use FP32 accumulation for reductions.
W16A16 is applied to SchNet MLP submodules, as described in Section~\ref{sec:w16a16}.

\subsection{Flash message passing fused edge computation}
\paragraph{Fused forward operator.}
For each edge $e$ with $(j,i)=(\mathtt{src}[e],\mathtt{dst}[e])$, we compute:
\begin{align*}
\mathbf{u}_e&=\mathbf{r}_i-\mathbf{r}_j,\quad
d_e=\|\mathbf{u}_e\|_2,\\
\tilde{\mathbf{b}}_e&=\mathrm{RBF}(d_e)\cdot C(d_e),\quad
\mathbf{w}_e=\mathrm{MLP}_{\mathrm{filter}}(\tilde{\mathbf{b}}_e),
\end{align*}
and then form the message $\mathbf{m}_e^{(t)}=\mathbf{x}_j^{(t)}\odot \mathbf{w}_e$ and directly feed it into aggregation for $\mathbf{h}_i^{(t)}$.
This removes the need to materialize $\mathtt{B}$ and $\mathtt{W}$ as HBM intermediates.

\paragraph{On-chip reuse.}
We tile edges and organize computation so that values reused within a short window, such as $\mathbf{r}_i$ for edges sharing the same destination, are kept in registers or shared memory.
This reduces global memory traffic even when it introduces modest recomputation.

\subsection{Flash aggregation segmented reductions}
\paragraph{Destination grouped segmented reduction in forward pass.}
To avoid atomic contention, we reorder edges by destination and perform a segmented reduction.
Let $\mathtt{dst\_ptr}\in\{0,\dots,E\}^{N+1}$ and $\mathtt{perm}\in\{1,\dots,E\}^{E}$ define a destination grouped layout where edges for node $i$ occupy as
\begin{align*}
p\in[\mathtt{dst\_ptr}[i],\,\mathtt{dst\_ptr}[i+1]).
\end{align*}
Then
\begin{align*}
\mathbf{h}_i^{(t)}=\sum_{p=\mathtt{dst\_ptr}[i]}^{\mathtt{dst\_ptr}[i+1]-1}
\mathbf{m}_{\mathtt{perm}[p]}^{(t)}.
\end{align*}
We assign exclusive ownership of each destination segment to one block, accumulate in registers, and write once per feature channel.

\paragraph{Source grouped segmented reduction in backward pass.}
The dominant gradient with respect to source features is as:
\begin{equation*}
\nabla \mathbf{x}_j^{(t)}=\sum_{e:\ \mathtt{src}[e]=j}
\nabla \mathbf{h}_{\mathtt{dst}[e]}^{(t)}\odot \mathbf{w}_e.
\end{equation*}
We avoid atomic contention by building a source grouped layout and applying the same exclusive ownership principle to accumulate $\nabla \mathbf{x}_j^{(t)}$.

\paragraph{Index construction under dynamic neighbor lists.}
Neighbor lists may change across MD steps, so the grouped layouts must be rebuilt when $(\mathtt{src},\mathtt{dst})$ changes.
We construct the destination grouped and source grouped indices using bucket sort on $\mathtt{dst}$ and $\mathtt{src}$, respectively, producing contiguous edge segments per node that enable exclusive ownership segmented reductions.
We report this bucket sort overhead jointly with the overall speedup in Section~\ref{sec:experiments}.

\subsection{W16A16 mixed precision for MLP submodules}
\label{sec:w16a16}

\begin{wrapfigure}{r}{0.55\linewidth}
\centering
\vspace{-10pt}
\includegraphics[width=\linewidth]{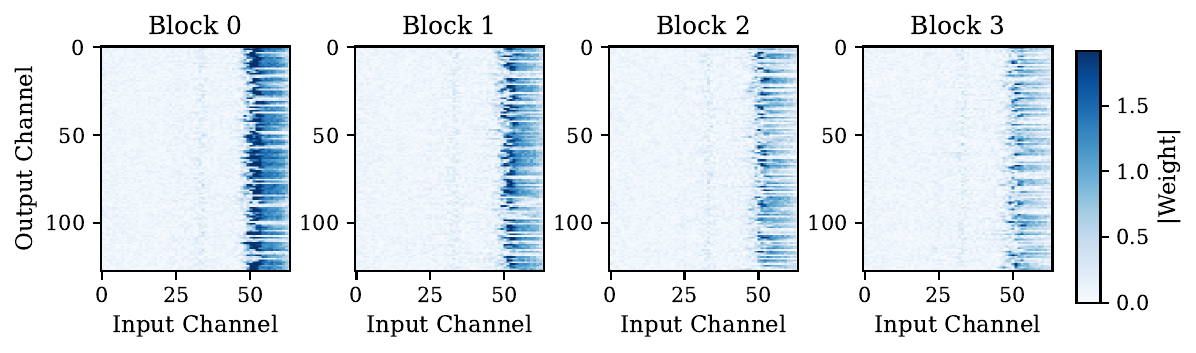}
\caption{Filter networks show clear channel-wise magnitude distribution, motivating channel quantization for lossless acceleration.}
\label{fig:weight-distribution}
\vspace{-10pt}
\end{wrapfigure}

\paragraph{Motivation.}
SchNet filter networks demonstrate a clear channel-wise magnitude structure.
As shown in Figure~\ref{fig:weight-distribution}, weight magnitudes concentrate unevenly across output channels, and this pattern is consistent across interaction blocks, motivating channel-wise quantization as a near-lossless way to reduce MLP computing and IO cost.

\paragraph{Channel-wise quantization.}
We apply W16A16~(16-bit weight, 16-bit activation) to all MLP submodules, including $\mathrm{MLP}_{\mathrm{filter}}$, blockwise update networks, and the readout network $\mathrm{MLP}_{\mathrm{out}}$.
We adapt Optimal Brain Compression~\citep{frantar2023optimalbraincompressionframework} to compute per-channel quantization scales for each linear layer, minimizing the quantization loss.
With FP16 weights and intermediate activations, MLP GEMMs map to Tensor Cores and reduce weight and activation traffic over FP32.
We keep positions $\mathbf{r}_i$, distances $d_e$, energy accumulation $\mathcal{E}$, and force outputs $\mathbf{F}_i$ in FP32, and use FP32 accumulation for reductions as described in Section~\ref{sec:io-aware-reform}.
\subsection{End-to-end integration}

At each MD step, we build the neighbor list, update segmented reduction indices when enabled, run $T$ interaction blocks with fused message passing and segmented reductions, compute energy via the readout network, and obtain forces by autodiff as $\mathbf{F}_i=-\nabla_{\mathbf{r}_i}\mathcal{E}$.
The configuration enables controlled ablations of fusion, segmented reductions, and W16A16 in Section~\ref{sec:experiments}.

\paragraph{Cost analysis.}
\texttt{FlashSchNet} avoids materializing $\mathtt{B}\in\mathbb{R}^{E\times D_r}$ and $\mathtt{W}\in\mathbb{R}^{E\times D}$ in HBM, reducing dominant IO per step from
$\mathrm{IO}^{\texttt{base}} = \Theta\!\big(T\cdot E (D_r + D)\big) + \Theta\!\big(T\cdot E D\big)$
to
$\mathrm{IO}^{\texttt{flash}} = \Theta\!\big(T\cdot(E D + N D)\big)$,
eliminating radial-basis and filter materialization, and replacing $O(ED)$ contention-heavy atomic aggregation with $O(ND)$ contention-free segment stores.
Since $E \gg N$ in typical GNN-MD (\textit{e.g.}, $10^5$ vs.\ $10^2$), total IO drops by $\sim\!E/N$.
16-bit quantization further reduces MLP weight and activation traffic by half.
\section{Empirical evaluations}
\label{sec:experiments}

\subsection{End-to-end results}
\paragraph{Experimental setting.}
We evaluate \texttt{FlashSchNet} on five fast-folding proteins following the benchmark suite of \citet{charron2025navigating}: Chignolin (CLN, 10 residues), TRPcage (2JOF, 20 residues), Homeodomain (1ENH, 54 residues), Villin (1YRF, 35 residues), and Alpha3D (2A3D, 73 residues). All simulations use Langevin dynamics at $300$~K with $64$ parallel replicas and the step size of $4$ fs on a single NVIDIA RTX PRO 6000 GPU. We compare against three baselines: CGSchNet~\cite{charron2025navigating} (the FP32 reference MLFF), the classical MARTINI force field~\cite{marrink2007martini}, and all-atom simulations. Structural fidelity is assessed via C$\alpha$ RMSD, fraction of native contacts $Q$, and GDT-TS. Throughput is reported in timestep$\cdot$mol/s (\textit{i.e.}, simulation steps per second aggregated over all replicas). More details are included in Appendix~\ref{app:metrics}.

\begin{figure}[ht]
\centering
\includegraphics[width=0.5\linewidth]{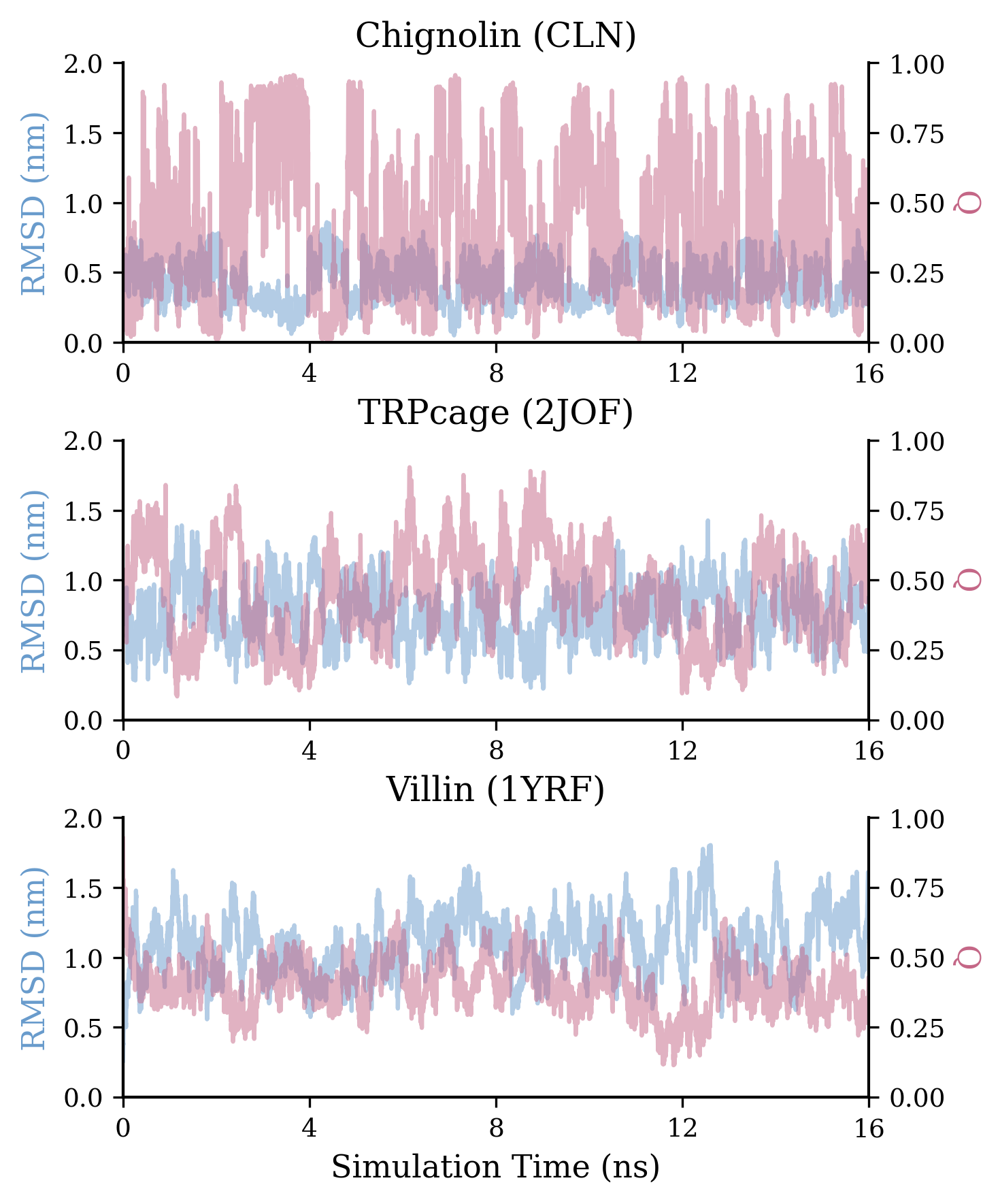}
\caption{Trajectories of C$\alpha$ RMSD and fraction of native contacts ($Q$) for three fast-folding proteins simulated with \texttt{FlashSchNet}. The plots demonstrate multiple reversible folding/unfolding events with the expected anti-correlation between RMSD and $Q$. \texttt{FlashSchNet} successfully captures the distinct folding timescales of Chignolin (nanosecond transitions) compared to the longer residence times of TRPcage and Villin.}
\label{fig:stability}
\end{figure}

\paragraph{Folding dynamics are preserved.}
To verify that our optimizations preserve the physical fidelity of the underlying potential, we simulate Chignolin, TRPcage, and Villin for 16~ns each. Figure~\ref{fig:stability} shows the evolution of RMSD and $Q$ over simulation time. All trajectories exhibit multiple reversible folding transitions with the expected anti-correlation between RMSD and $Q$. Chignolin shows rapid nanosecond-scale transitions between folded ($Q > 0.8$) and unfolded ($Q < 0.4$) states; TRPcage exhibits dynamic fluctuations with $Q$ oscillating between 0.4 and 0.9; Villin displays longer residence times in metastable basins, reaching the native state ($Q > 0.85$) multiple times. These results confirm that \texttt{FlashSchNet} correctly samples the conformational landscape without introducing numerical artifacts.

\begin{table}[h]
\centering
\caption{Structural accuracy benchmark. \texttt{FlashSchNet} retains the high fidelity of the baseline CGSchNet and substantially outperforms the classical MARTINI model. All-Atom simulations serve as the experimental reference.}
\label{tab:metrics}
\resizebox{0.68\linewidth}{!}{
\begin{tabular}{ll|cc|c|c}
\toprule[1pt]
\midrule
\multirow{2}{*}{\textbf{Protein}} & \multirow{2}{*}{\textbf{Metric}} & \multicolumn{2}{c|}{\textbf{MLFFs}} & \textbf{Classical} & \textbf{Reference} \\
& & \texttt{FlashSchNet} & CGSchNet & MARTINI & All-Atom \\
\midrule
\multirow{2}{*}{Chignolin} & GDT-TS & 0.90 & 0.90 & 0.66 & 1.00 \\
 & Largest $Q$ & 0.89 & 0.96 & 0.83 & 0.95 \\
\midrule
\multirow{2}{*}{TRPcage} & GDT-TS & 0.72 & 0.72 & 0.64 & 0.88 \\
 & Largest $Q$ & 0.89 & 0.96 & 0.60 & 0.95 \\
\midrule
\multirow{2}{*}{Villin} & GDT-TS & 0.74 & 0.78 & 0.46 & 0.88 \\
 & Largest $Q$ & 0.88 & 0.96 & 0.56 & 0.93 \\
\midrule
\bottomrule[1pt]
\end{tabular}
}
\end{table}
\paragraph{Structural fidelity matches CGSchNet baseline.}
Table~\ref{tab:metrics} benchmarks structural accuracy using GDT-TS and the largest metastable $Q$. \texttt{FlashSchNet} maintains GDT-TS scores within 0.04 of the CGSchNet baseline across all proteins, while both MLFFs substantially outperform MARTINI in stabilizing near-native structures. A similar trend holds for the largest metastable $Q$ that \texttt{FlashSchNet} consistently reaches $Q \geq 0.88$, comparable to CGSchNet and significantly higher than MARTINI ($Q \approx 0.56$--$0.83$). These findings confirm that \texttt{FlashSchNet} preserves the physical accuracy of the original CGSchNet model while improving simulation speed significantly.

\begin{table}[h]
\centering
\caption{Computational efficiency benchmark. Evaluated proteins include Chignolin (CLN), TRPcage (2JOF), Homeodomain (1ENH), Villin (1YRF), and Alpha3D (2A3D). Performance metrics reported are speed (timestep$\cdot$mol/s) and peak memory (GB). \texttt{FlashSchNet} demonstrates competitive throughput compared to classical benchmarks on a single RTX PRO 6000 GPU.}
\label{tab:performance_filled}
\resizebox{0.78\linewidth}{!}{
\begin{tabular}{ll|cc|c|c}
\toprule[1pt]
\midrule
\multirow{2}{*}{\textbf{Protein system}} & \multirow{2}{*}{\textbf{Metric}} & \multicolumn{2}{c|}{\textbf{MLFF}} & \textbf{Classical} & \textbf{Reference} \\
& & \texttt{FlashSchNet} & CGSchNet & MARTINI & All-Atom \\
\midrule
\multirow{2}{*}{Chignolin} & Speed & $\mathbf{5222}$ & $3578$ & $2580$ & $1437$ \\
 & Peak Mem. & $\mathbf{3.7}$ & $22.7$ & $35.0$ & $36.3$ \\
\midrule
\multirow{2}{*}{TRPcage} & Speed & $\mathbf{4938}$ & $1729$ & $2550$ & $1419$ \\
 & Peak Mem. & $\mathbf{8.8}$ & $29.2$ & $34.9$ & $38.3$ \\
\midrule
\multirow{2}{*}{Homeodomain} & Speed & $\mathbf{3095}$ & $477$ & $2250$ & $1005$ \\
 & Peak Mem. & $\mathbf{18.0}$ & $92.5$ & $34.9$ & $37.4$ \\
\midrule
\multirow{2}{*}{Villin} & Speed & $\mathbf{3912}$ & $1056$ & $2340$ & $1275$ \\
 & Peak Mem. & $\mathbf{12.9}$ & $94.2$ & $35.0$ & $47.9$ \\
\midrule
\multirow{2}{*}{Alpha3D} & Speed & $\mathbf{2610}$ & $288$ & $2160$ & $861$ \\
 & Peak Mem. & $\mathbf{22.4}$ & $94.1$ & $31.7$ & $63.6$ \\
\midrule
\bottomrule[1pt]
\end{tabular}
}
\end{table}

\paragraph{Throughput reaches classical force field parity.}
Table~\ref{tab:performance_filled} summarizes throughput and memory usage. On the Homeodomain (1ENH) system, \texttt{FlashSchNet} achieves around \textbf{$\mathbf{3000}$ timestep$\cdot$mol/s}~(\textit{i.e.} 1000 ns/day), a \textbf{$\mathbf{6.5\times}$ speedup} over the CGSchNet baseline (around $500$ timestep$\cdot$mol/s). This effectively closes the gap between MLFFs and classical potentials, as \texttt{FlashSchNet} reaches parity with MARTINI (around $2900$ timestep$\cdot$mol/s) while significantly outperforming all-atom simulations (around 1200 timestep$\cdot$mol/s).
Moreover, \texttt{FlashSchNet} reduces peak memory from 92GB (CGSchNet) to \textbf{18GB} (\textbf{$>80\%$ reduction}) by removing materialization of large intermediates. This potentially enables simulations of large systems on commodity hardware (\textit{e.g.} a single RTX 5090).

\subsection{Ablation results}

\begin{figure}[htbp]
\centering
\includegraphics[width=0.75\linewidth]{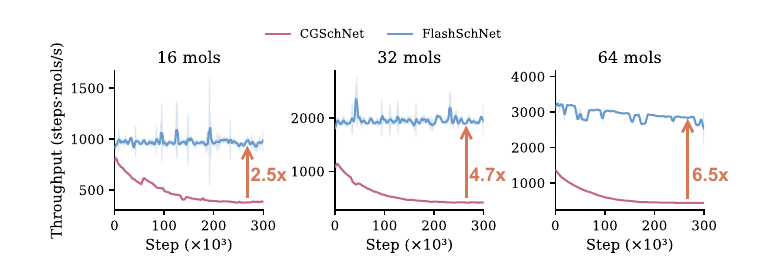}
\caption{Step-wise throughput comparison on 1ENH protein during 300k-step elongated simulation across three batch sizes (\textit{i.e.} $16$, $32$, $64$ parallel replicas). \texttt{FlashSchNet} maintains consistent throughput along simulation despite evolving graph topology, while CGSchNet degrades as the neighbor graph becomes denser and less diagonal, as shown in Figure~\ref{fig:periodic-graph}. The speedup gap widens with batch size, reaching $6.5\times$ at $64$ replicas.}
\label{fig:periodic-throughput}
\end{figure}

\begin{figure}[htbp]
\centering
\includegraphics[width=0.75\linewidth]{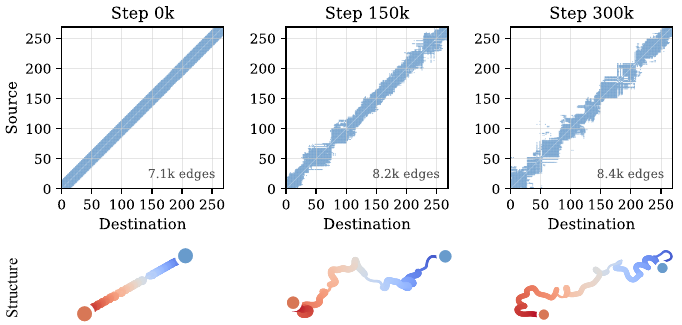}
\caption{Evolution of graph topology and protein structure during 1ENH elongated simulation. \uline{Top:} Adjacency matrices at steps $0$, $150$k, and $300$k, showing increasing off-diagonal density as the protein unfolds (\textit{e.g.} edges grow from $7.1$k to $8.4$k). \uline{Bottom:} Corresponding 3D structures colored by residue index, illustrating the transition from compact folded state to extended conformations.}
\label{fig:periodic-graph}
\end{figure}

\paragraph{Robustness to dynamic graph topology.}
A key challenge in GNN-MD is that neighbor graphs evolve throughout simulation, particularly during conformational transitions. As shown in Figure~\ref{fig:periodic-graph}, the elongated 1ENH protein unfolds over $300$k steps, causing the adjacency matrix to shift from near-diagonal to dense off-diagonal structure, with edge count increasing. Figure~\ref{fig:periodic-throughput} reveals that CGSchNet throughput degrades substantially under these conditions, likely due to increased scatter contention when edges distribute across more destination nodes~\citep{gong2025identifying}. In contrast, \texttt{FlashSchNet} maintains stable throughput via contention-free CSR segment reductions, which are agnostic to edge distribution patterns. This robustness is critical for practical MD workflows involving large conformational changes.

\begin{figure}[t]
\centering
\includegraphics[width=0.7\linewidth]{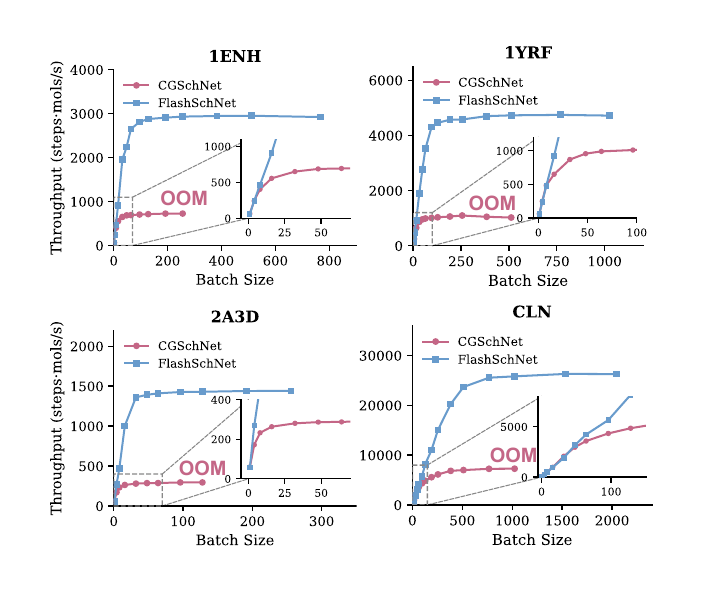}
\caption{Throughput scaling with batch size across four protein systems on a single RTX PRO 6000 GPU. \texttt{FlashSchNet} consistently scales to $3$-$10\times$ larger batch sizes than CGSchNet before saturating, while CGSchNet encounters out-of-memory~(OOM) at much smaller batch sizes. The memory reduction from our IO-aware design is critical for enhanced sampling workflows requiring many parallel trajectories.}
\label{fig:throughput-scaling}
\end{figure}

\paragraph{Memory reduction enables better scalability.}
As shown in Figure~\ref{fig:throughput-scaling}, we examine how throughput scales with the number of parallel replicas across four protein systems of varying size: Chignolin (CLN, 10 residues), Villin (1YRF, 35 residues), Homeodomain (1ENH, 54 residues), and Alpha3D (2A3D, 73 residues). CGSchNet exhausts GPU memory at small batch sizes across all systems (see insets), limiting its utility for enhanced sampling methods such as replica exchange that benefit from many concurrent trajectories. In contrast, \texttt{FlashSchNet} scales to $3$-$10\times$ larger batch sizes depending on system size, \textit{i.e.} from $256$ replicas for the largest protein (Alpha3D) to $2048$ replicas for Chignolin. Throughput grows near-linearly before gradually saturating as compute resources become fully utilized. This scalability is particularly valuable for workflows requiring statistical convergence over many independent trajectories.
\section{Conclusion}
\label{sec:conclusion}
We present \texttt{FlashSchNet}, an IO-aware SchNet-style GNN molecular dynamics framework that addresses the memory-bound nature of learned potentials.
By identifying HBM traffic as the key bottleneck rather than FLOPs, we developed four techniques that exploit inherent model structure to reduce data movement at the algorithmic level.
\textit{Flash radial basis} fuses distance computation and basis expansion into a single tiled pass.
\textit{Flash message passing} eliminates intermediate edge tensor materialization.
\textit{Flash aggregation} reformulates scatter-add via CSR segment reduce for contention-free accumulation.
\textit{Channel-wise 16-bit quantization} exploits low per-channel dynamic range to further improve throughput.
Together, these techniques achieve \textbf{6.5$\times$ speedup} and \textbf{80\% memory reduction} over the CGSchNet baseline, reaching 1000 ns/day aggregate throughput on coarse-grained protein containing 269 beads across 64 parallel replicas on a single RTX PRO 6000 GPU.
To our knowledge, \texttt{FlashSchNet} is the first SchNet-style GNN-MD that is faster than classical coarse-grained force fields, \textit{e.g.} MARTINI, in wall-clock efficiency while retaining the accuracy and transferability of learned potentials.

\section*{Impact Statement}
\texttt{FlashSchNet} improves the efficiency and memory footprint of SchNet-style GNN molecular dynamics through IO-aware fused kernels, contention-free aggregation, and lightweight quantization. These gains enable more concurrent multi-replica simulations on a fixed GPU budget, improving statistical efficiency and coverage of rare events. This can broaden access to accurate learned MD for academic and industrial users in computational chemistry, drug discovery, and materials science. By increasing utilization and reducing redundant memory movement, the techniques may also lower energy per simulated nanosecond, although net environmental impact depends on whether efficiency gains lead to more total simulation.

\bibliography{999_reference}

@article{karplus2002molecular,
  title={Molecular dynamics simulations of biomolecules},
  author={Karplus, Martin and McCammon, J Andrew},
  journal={Nature structural biology},
  volume={9},
  number={9},
  pages={646--652},
  year={2002},
  publisher={Nature Publishing Group US New York}
}

@article{hollingsworth2018molecular,
  title={Molecular dynamics simulation for all},
  author={Hollingsworth, Scott A and Dror, Ron O},
  journal={Neuron},
  volume={99},
  number={6},
  pages={1129--1143},
  year={2018},
  publisher={Elsevier}
}

@article{car1985unified,
  title={Unified approach for molecular dynamics and density-functional theory},
  author={Car, Richard and Parrinello, Mark},
  journal={Physical review letters},
  volume={55},
  number={22},
  pages={2471},
  year={1985},
  publisher={APS}
}

@article{kuhne2020cp2k,
  title={CP2K: An electronic structure and molecular dynamics software package-Quickstep: Efficient and accurate electronic structure calculations},
  author={K{\"u}hne, Thomas D and Iannuzzi, Marcella and Del Ben, Mauro and Rybkin, Vladimir V and Seewald, Patrick and Stein, Frederick and Laino, Teodoro and Khaliullin, Rustam Z and Sch{\"u}tt, Ole and Schiffmann, Florian and others},
  journal={The Journal of Chemical Physics},
  volume={152},
  number={19},
  year={2020},
  publisher={AIP Publishing}
}

@article{wang2004development,
  title={Development and testing of a general amber force field},
  author={Wang, Junmei and Wolf, Romain M and Caldwell, James W and Kollman, Peter A and Case, David A},
  journal={Journal of computational chemistry},
  volume={25},
  number={9},
  pages={1157--1174},
  year={2004},
  publisher={Wiley Online Library}
}

@article{brooks2009charmm,
  title={CHARMM: the biomolecular simulation program},
  author={Brooks, Bernard R and Brooks III, Charles L and Mackerell Jr, Alexander D and Nilsson, Lennart and Petrella, Robert J and Roux, Beno{\^\i}t and Won, Youngdo and Archontis, Georgios and Bartels, Christian and Boresch, Stefan and others},
  journal={Journal of computational chemistry},
  volume={30},
  number={10},
  pages={1545--1614},
  year={2009},
  publisher={Wiley Online Library}
}

@article{bartok2010gaussian,
  title={Gaussian approximation potentials: The accuracy of quantum mechanics, without the electrons},
  author={Bart{\'o}k, Albert P and Payne, Mike C and Kondor, Risi and Cs{\'a}nyi, G{\'a}bor},
  journal={Physical review letters},
  volume={104},
  number={13},
  pages={136403},
  year={2010},
  publisher={APS}
}

@article{chmiela2019sgdml,
  title={sGDML: Constructing accurate and data efficient molecular force fields using machine learning},
  author={Chmiela, Stefan and Sauceda, Huziel E and Poltavsky, Igor and M{\"u}ller, Klaus-Robert and Tkatchenko, Alexandre},
  journal={Computer Physics Communications},
  volume={240},
  pages={38--45},
  year={2019},
  publisher={Elsevier}
}

@article{behler2007generalized,
  title={Generalized neural-network representation of high-dimensional potential-energy surfaces},
  author={Behler, J{\"o}rg and Parrinello, Michele},
  journal={Physical review letters},
  volume={98},
  number={14},
  pages={146401},
  year={2007},
  publisher={APS}
}

@article{schutt2017schnet,
  title={Schnet: A continuous-filter convolutional neural network for modeling quantum interactions},
  author={Sch{\"u}tt, Kristof and Kindermans, Pieter-Jan and Sauceda Felix, Huziel Enoc and Chmiela, Stefan and Tkatchenko, Alexandre and M{\"u}ller, Klaus-Robert},
  journal={Advances in neural information processing systems},
  volume={30},
  year={2017}
}

@article{gasteiger2020directional,
  title={Directional message passing for molecular graphs},
  author={Gasteiger, Johannes and Gro{\ss}, Janek and G{\"u}nnemann, Stephan},
  journal={arXiv preprint arXiv:2003.03123},
  year={2020}
}

@article{unke2019physnet,
  title={PhysNet: A neural network for predicting energies, forces, dipole moments, and partial charges},
  author={Unke, Oliver T and Meuwly, Markus},
  journal={Journal of chemical theory and computation},
  volume={15},
  number={6},
  pages={3678--3693},
  year={2019},
  publisher={ACS Publications}
}

@article{batzner20223,
  title={E (3)-equivariant graph neural networks for data-efficient and accurate interatomic potentials},
  author={Batzner, Simon and Musaelian, Albert and Sun, Lixin and Geiger, Mario and Mailoa, Jonathan P and Kornbluth, Mordechai and Molinari, Nicola and Smidt, Tess E and Kozinsky, Boris},
  journal={Nature communications},
  volume={13},
  number={1},
  pages={2453},
  year={2022},
  publisher={Nature Publishing Group UK London}
}

@article{musaelian2023learning,
  title={Learning local equivariant representations for large-scale atomistic dynamics},
  author={Musaelian, Albert and Batzner, Simon and Johansson, Anders and Sun, Lixin and Owen, Cameron J and Kornbluth, Mordechai and Kozinsky, Boris},
  journal={Nature Communications},
  volume={14},
  number={1},
  pages={579},
  year={2023},
  publisher={Nature Publishing Group UK London}
}

@article{batatia2022mace,
  title={MACE: Higher order equivariant message passing neural networks for fast and accurate force fields},
  author={Batatia, Ilyes and Kovacs, David P and Simm, Gregor and Ortner, Christoph and Cs{\'a}nyi, G{\'a}bor},
  journal={Advances in neural information processing systems},
  volume={35},
  pages={11423--11436},
  year={2022}
}

@article{ju2025application,
  title={Application of pretrained universal machine-learning interatomic potential for physicochemical simulation of liquid electrolytes in Li-ion batteries},
  author={Ju, Suyeon and You, Jinmu and Kim, Gijin and Park, Yutack and An, Hyungmin and Han, Seungwu},
  journal={Digital Discovery},
  year={2025},
  publisher={Royal Society of Chemistry}
}

@article{neumann2024orb,
  title={Orb: A fast, scalable neural network potential},
  author={Neumann, Mark and Gin, James and Rhodes, Benjamin and Bennett, Steven and Li, Zhiyi and Choubisa, Hitarth and Hussey, Arthur and Godwin, Jonathan},
  journal={arXiv preprint arXiv:2410.22570},
  year={2024}
}

@article{yang2024mattersim,
  title={Mattersim: A deep learning atomistic model across elements, temperatures and pressures},
  author={Yang, Han and Hu, Chenxi and Zhou, Yichi and Liu, Xixian and Shi, Yu and Li, Jielan and Li, Guanzhi and Chen, Zekun and Chen, Shuizhou and Zeni, Claudio and others},
  journal={arXiv preprint arXiv:2405.04967},
  year={2024}
}

@article{charron2025navigating,
  title={Navigating protein landscapes with a machine-learned transferable coarse-grained model},
  author={Charron, Nicholas E and Bonneau, Klara and Pasos-Trejo, Aldo S and Guljas, Andrea and Chen, Yaoyi and Musil, F{\'e}lix and Venturin, Jacopo and Gusew, Daria and Zaporozhets, Iryna and Kr{\"a}mer, Andreas and others},
  journal={Nature chemistry},
  volume={17},
  number={8},
  pages={1284--1292},
  year={2025},
  publisher={Nature Publishing Group UK London}
}

@article{majewski2023machine,
  title={Machine learning coarse-grained potentials of protein thermodynamics},
  author={Majewski, Maciej and P{\'e}rez, Adri{\`a} and Th{\"o}lke, Philipp and Doerr, Stefan and Charron, Nicholas E and Giorgino, Toni and Husic, Brooke E and Clementi, Cecilia and No{\'e}, Frank and De Fabritiis, Gianni},
  journal={Nature Communications},
  volume={14},
  number={1},
  pages={5739},
  year={2023},
  publisher={Nature Publishing Group UK London}
}

@article{airas2026knowledge,
  title={Knowledge Distillation of a Protein Language Model Yields a Foundational Implicit Solvent Model},
  author={Airas, Justin and Zhang, Bin},
  journal={arXiv preprint arXiv:2601.05388},
  year={2026}
}

@article{dror2012biomolecular,
  title={Biomolecular simulation: a computational microscope for molecular biology},
  author={Dror, Ron O and Dirks, Robert M and Grossman, JP and Xu, Huafeng and Shaw, David E},
  journal={Annual review of biophysics},
  volume={41},
  pages={429--452},
  year={2012},
  publisher={Annual Reviews}
}

@article{de2016role,
  title={Role of molecular dynamics and related methods in drug discovery},
  author={De Vivo, Marco and Masetti, Matteo and Bottegoni, Giovanni and Cavalli, Andrea},
  journal={Journal of medicinal chemistry},
  volume={59},
  number={9},
  pages={4035--4061},
  year={2016},
  publisher={ACS Publications}
}

@article{chodera2014markov,
  title={Markov state models of biomolecular conformational dynamics},
  author={Chodera, John D and No{\'e}, Frank},
  journal={Current opinion in structural biology},
  volume={25},
  pages={135--144},
  year={2014},
  publisher={Elsevier}
}

@article{marrink2007martini,
  title={The MARTINI force field: coarse grained model for biomolecular simulations},
  author={Marrink, Siewert J and Risselada, H Jelger and Yefimov, Serge and Tieleman, D Peter and De Vries, Alex H},
  journal={The journal of physical chemistry B},
  volume={111},
  number={27},
  pages={7812--7824},
  year={2007},
  publisher={ACS Publications}
}

@article{bronstein2021geometric,
  title={Geometric deep learning: Grids, groups, graphs, geodesics, and gauges},
  author={Bronstein, Michael M and Bruna, Joan and Cohen, Taco and Veli{\v{c}}kovi{\'c}, Petar},
  journal={arXiv preprint arXiv:2104.13478},
  year={2021}
}

@inproceedings{gilmer2017neural,
  title={Neural message passing for quantum chemistry},
  author={Gilmer, Justin and Schoenholz, Samuel S and Riley, Patrick F and Vinyals, Oriol and Dahl, George E},
  booktitle={International conference on machine learning},
  pages={1263--1272},
  year={2017},
  organization={Pmlr}
}

@article{wang2019deep,
  title={Deep graph library: A graph-centric, highly-performant package for graph neural networks},
  author={Wang, Minjie and Zheng, Da and Ye, Zihao and Gan, Quan and Li, Mufei and Song, Xiang and Zhou, Jinjing and Ma, Chao and Yu, Lingfan and Gai, Yu and others},
  journal={arXiv preprint arXiv:1909.01315},
  year={2019}
}

@article{fey2019fast,
  title={Fast graph representation learning with PyTorch Geometric},
  author={Fey, Matthias and Lenssen, Jan Eric},
  journal={arXiv preprint arXiv:1903.02428},
  year={2019}
}

@inproceedings{wang2021gnnadvisor,
  title={$\{$GNNAdvisor$\}$: An adaptive and efficient runtime system for $\{$GNN$\}$ acceleration on $\{$GPUs$\}$},
  author={Wang, Yuke and Feng, Boyuan and Li, Gushu and Li, Shuangchen and Deng, Lei and Xie, Yuan and Ding, Yufei},
  booktitle={15th USENIX symposium on operating systems design and implementation (OSDI 21)},
  pages={515--531},
  year={2021}
}

@article{xie2022graphiler,
  title={Graphiler: Optimizing graph neural networks with message passing data flow graph},
  author={Xie, Zhiqiang and Wang, Minjie and Ye, Zihao and Zhang, Zheng and Fan, Rui},
  journal={Proceedings of Machine Learning and Systems},
  volume={4},
  pages={515--528},
  year={2022}
}

@inproceedings{chen2020fusegnn,
  title={fuseGNN: Accelerating graph convolutional neural network training on GPGPU},
  author={Chen, Zhaodong and Yan, Mingyu and Zhu, Maohua and Deng, Lei and Li, Guoqi and Li, Shuangchen and Xie, Yuan},
  booktitle={Proceedings of the 39th International Conference on Computer-Aided Design},
  pages={1--9},
  year={2020}
}

@inproceedings{huang2020ge,
  title={Ge-spmm: General-purpose sparse matrix-matrix multiplication on gpus for graph neural networks},
  author={Huang, Guyue and Dai, Guohao and Wang, Yu and Yang, Huazhong},
  booktitle={SC20: International Conference for High Performance Computing, Networking, Storage and Analysis},
  pages={1--12},
  year={2020},
  organization={IEEE}
}

@inproceedings{rahman2021fusedmm,
  title={Fusedmm: A unified sddmm-spmm kernel for graph embedding and graph neural networks},
  author={Rahman, Md Khaledur and Sujon, Majedul Haque and Azad, Ariful},
  booktitle={2021 IEEE International Parallel and Distributed Processing Symposium (IPDPS)},
  pages={256--266},
  year={2021},
  organization={IEEE}
}

@article{dao2022flashattention,
  title={Flashattention: Fast and memory-efficient exact attention with io-awareness},
  author={Dao, Tri and Fu, Dan and Ermon, Stefano and Rudra, Atri and R{\'e}, Christopher},
  journal={Advances in neural information processing systems},
  volume={35},
  pages={16344--16359},
  year={2022}
}

@article{pelaez2024torchmd,
  title={Torchmd-net 2.0: Fast neural network potentials for molecular simulations},
  author={Pelaez, Raul P and Simeon, Guillem and Galvelis, Raimondas and Mirarchi, Antonio and Eastman, Peter and Doerr, Stefan and Tholke, Philipp and Markland, Thomas E and De Fabritiis, Gianni},
  journal={Journal of Chemical Theory and Computation},
  volume={20},
  number={10},
  pages={4076--4087},
  year={2024},
  publisher={ACS Publications}
}

@inproceedings{gong2025identifying,
  title={Identifying and Analyzing Pitfalls in $\{$GNN$\}$ Systems},
  author={Gong, Yidong and Tarafder, Arnab Kanti and Afrin, Saima and Kumar, Pradeep},
  booktitle={2025 USENIX Annual Technical Conference (USENIX ATC 25)},
  pages={1605--1624},
  year={2025}
}

@article{liu2024df,
  title={DF-GNN: Dynamic Fusion Framework for Attention Graph Neural Networks on GPUs},
  author={Liu, Jiahui and Cai, Zhenkun and Chen, Zhiyong and Wang, Minjie},
  journal={arXiv preprint arXiv:2411.16127},
  year={2024}
}

@misc{frantar2023optimalbraincompressionframework,
      title={Optimal Brain Compression: A Framework for Accurate Post-Training Quantization and Pruning},
      author={Elias Frantar and Sidak Pal Singh and Dan Alistarh},
      year={2023},
      eprint={2208.11580},
      archivePrefix={arXiv},
      primaryClass={cs.LG},
      url={https://arxiv.org/abs/2208.11580},
}

@article{zemla2003lga,
  title={LGA: a method for finding 3D similarities in protein structures},
  author={Zemla, Adam},
  journal={Nucleic acids research},
  volume={31},
  number={13},
  pages={3370--3374},
  year={2003},
  publisher={Oxford University Press}
}

@article{zhang2004scoring,
  title={Scoring function for automated assessment of protein structure template quality},
  author={Zhang, Yang and Skolnick, Jeffrey},
  journal={Proteins: Structure, Function, and Bioinformatics},
  volume={57},
  number={4},
  pages={702--710},
  year={2004},
  publisher={Wiley Online Library}
}

@article{best2013native,
  title={Native contacts determine protein folding mechanisms in atomistic simulations},
  author={Best, Robert B and Hummer, Gerhard and Eaton, William A},
  journal={Proceedings of the National Academy of Sciences},
  volume={110},
  number={44},
  pages={17874--17879},
  year={2013},
  publisher={National Academy of Sciences}
}
\bibliographystyle{style/icml2025}

\titlespacing*{\section}{0pt}{*1}{*1}
\titlespacing*{\subsection}{0pt}{*1.25}{*1.25}
\titlespacing*{\subsubsection}{0pt}{*1.5}{*1.5}

\setlength{\abovedisplayskip}{\baselineskip} 
\setlength{\abovedisplayshortskip}{0.5\baselineskip} 
\setlength{\belowdisplayskip}{\baselineskip}
\setlength{\belowdisplayshortskip}{0.5\baselineskip}

\appendix

\appcontents

\newpage

\section{Details of evaluation metrics and protocols}
\label{app:metrics}

\subsection{Root Mean Square Deviation (RMSD)}

The C$\alpha$ Root Mean Square Deviation (RMSD) quantifies the geometric deviation between a sampled conformation and a reference structure. Given $N$ C$\alpha$ atom positions $\{\mathbf{r}_i\}_{i=1}^{N}$ in the query structure and corresponding positions $\{\mathbf{r}_i^{\text{ref}}\}_{i=1}^{N}$ in the reference, the RMSD is defined as the minimum Euclidean distance achievable under rigid-body transformation:

\begin{equation}
\text{RMSD} = \min_{\mathbf{R} \in \text{SO}(3), \mathbf{t} \in \mathbb{R}^3} \sqrt{\frac{1}{N} \sum_{i=1}^{N} \|\mathbf{R}\mathbf{r}_i + \mathbf{t} - \mathbf{r}_i^{\text{ref}}\|^2}
\end{equation}

where $\mathbf{R}$ denotes the rotation matrix and $\mathbf{t}$ the translation vector that optimally align the two structures. This superposition is typically computed via singular value decomposition (SVD). Lower RMSD values indicate higher structural fidelity to the native state.
\subsection{Fraction of Native Contacts ($Q$)}

The fraction of native contacts ($Q$) serves as a reaction coordinate to quantify structural similarity based on pairwise residue distances, capturing topological fidelity rather than global superposition~\cite{best2013native}. It is defined as:
\begin{equation}
Q = \frac{1}{N_c} \sum_{(i,j) \in \mathcal{C}} \frac{1}{1 + \exp\left[\beta(r_{ij} - \lambda r_{ij}^0)\right]}
\end{equation}
where $\mathcal{C}$ denotes the set of native contact pairs, $N_c = |\mathcal{C}|$ is the total number of native contacts, $r_{ij}$ is the C$\alpha$ distance between residues $i$ and $j$ in the query structure, and $r_{ij}^0$ is the corresponding distance in the reference native structure.

\paragraph{Native contact definition.}
We define a native contact for any residue pair $(i, j)$ that satisfies two criteria in the reference all-atom structure: (1) a sequence separation $|i - j| \geq 3$, and (2) a heavy-atom distance less than 4.5~\AA{}. The C$\alpha$ distances of these identified pairs form the reference set $\{r_{ij}^0\}$.

\paragraph{Parameters.}
Following~\cite{charron2025navigating}, we set $\beta = 10$~nm$^{-1}$ and $\lambda = 1.5$. The parameter $\beta$ modulates the steepness of the sigmoid function, while $\lambda$ accounts for thermal fluctuations around the native distance. These hyperparameters produce smooth free energy surfaces that clearly distinguish native-like states ($Q \approx 1$) from unfolded configurations ($Q < 0.5$).

\paragraph{Largest metastable $Q$.}
During simulation, the protein samples a probability distribution over $Q$. To evaluate the force field's ability to stabilize the native state, we compute the 1D probability density of $Q$, apply Savitzky-Golay smoothing, and identify the \textit{rightmost} local maximum (i.e., the stable basin with the highest $Q$ value). This metric indicates the structural fidelity of the folded state populated by the model.
\subsection{GDT-TS Score}

The Global Distance Test Total Score (GDT-TS)~\cite{zemla2003lga} quantifies structural similarity by identifying the maximal subset of C$\alpha$ atoms that can be superimposed within a defined distance cutoff. For a specific cutoff $d$, let $P_d$ denote the percentage of C$\alpha$ atoms in the query structure falling within $d$~\AA{} of their corresponding positions in the reference structure after optimal superposition. The GDT-TS is calculated as:
\begin{equation}
\text{GDT-TS} = \frac{P_1 + P_2 + P_4 + P_8}{4}
\end{equation}
where $P_1$, $P_2$, $P_4$, and $P_8$ correspond to cutoffs of 1, 2, 4, and 8~\AA{}, respectively. Unlike RMSD, GDT-TS is less sensitive to local high-variance regions (e.g., loops) and provides a robust metric for global topology. All GDT-TS calculations are performed using the TM-score program~\cite{zhang2004scoring}.

\paragraph{Evaluation protocol.}
Following the protocol in~\cite{charron2025navigating}, we construct a 2D free energy surface in RMSD vs.\ $Q$ space and apply $k$-means clustering with 100 centers. We identify the most native-like cluster (defined by the highest $Q$ and lowest RMSD) and randomly sample 10 representative structures from this basin. The reported GDT-TS is the average score of these samples against the experimental reference structure.

\end{document}